# Feature Engineering for Map Matching of Low-Sampling-Rate GPS Trajectories in Road Network

Jian Yang and Liqiu Meng

Lehrstuhl für Kartographie, Technische Universität München, 80333 Munich, Germany
{jian.yang@tum.de, liqiu.meng@bv.tu-muenchen.de}

**Abstract.** Map matching of GPS trajectories from a sequence of noisy observations serves the purpose of recovering the original routes in a road network. In this work in progress, we attempt to share our experience of feature construction in a spatial database by reporting our ongoing experiment of feature extraction in Conditional Random Fields (CRFs) for map matching. Our preliminary results are obtained from real-world taxi GPS trajectories.

Keywords. Map Matching, Feature Engineering, CRFs, Spatial Database

## 1 Introduction

Map Matching of GPS trajectories serves the purpose of recovering the original route on a road network from a sequence of GPS observations (see Fig.1). It is a fundamental technique for many Location Based Services (LBS) and has raised a lot of interest in recent years [1][2][3]. Many researches have achieved satisfying results of matching GPS trajectories at a moderate sampling rate [2]. However, the matching at a low sampling rate (sample interval longer than 120 seconds) remains a challenging task.

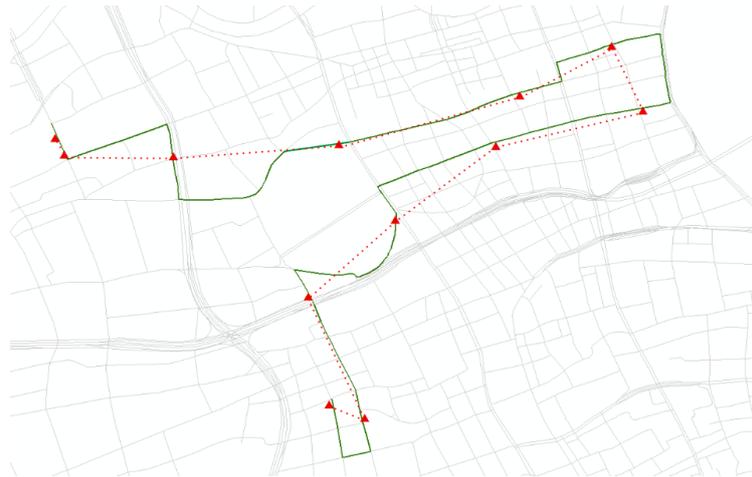

**Fig. 1.** GPS observations (red triangles) and the original driving route (green line) in a road network.

Map matching is often modeled as sequence labeling problem and can be solved with a Hidden Markov Model (HMM) [4] and its discriminating variant, Conditional Random Fields (CRFs) [5]. These methods sharing a similar probabilistic modeling framework but differ in the ways how they model the emission and transition probability, or they use different features. In order to improve the accuracy of map matching and reveal the key issues in modeling the uncertainty of trajectory data in a road network, we further investigate map matching problem from a feature engineering perspective.

In this work in progress, we report our ongoing experiment on feature extraction in CRFs for map matching. More specifically, we describe probabilistic modeling with CRFs and discuss feature construction with a spatial database. Our preliminary results in comparison to baseline methods are obtained on a sample dataset of real-world taxi GPS trajectories.

## 2 Map Matching with Conditional Random Fields

### 2.1 Map Matching

Map matching is characterized by a two-fold challenge: 1) Observations are often noisy due to the inaccurate GPS sensor or poor positioning conditions, e.g., low-speed maneuvers of vehicle in traffic, passage through urban canyons and tunnels. 2) A low sampling rate for data collection is often used to reduce power consumption and communication cost. This may cause an information loss between neighboring observations, complicating the route recovery as a large number of alternative paths can be found in the road network.

Map matching has invoked a growing interest in the recent years for its importance in LBS applications. A comprehensive literature survey was conducted in [6], in which map matching methods are categorized into four groups: geometric, topological, probabilistic, and other advanced techniques. Among these approaches, the probabilistic methods based on the Hidden Markov Model (HMM) are most popular because of its well-studied theoretical base and competitive performances [2][4][7][8]. A HMM-based method models the emission probability and transition probability from a GPS observation sequence with the aforementioned challenges of noisy measurement and insufficient sampling rate. HMM's discriminating variant, Conditional Random Fields (CRFs), is also applied to map matching successfully in a real-world case [5]. Thus, we further explore the use of CRFs for map matching of low-sampling-rate GPS trajectories.

### 2.2 A Chain Structured CRFs for Map Matching

The Conditional Random Fields (CRFs) is an undirected graphical model used to compute probability of a possible label sequence conditioned on the observation sequence [9]. The CRFs represents the conditional probability as the product of poten-tial functions over cliques in the graph, which are computed in terms of feature functions of random variables in the observation and label sequence.

We define two types of random variables to model the uncertainty in the sequence of GPS observations in an alternating order. Let $X = \{x_1, ..., x_N\}$ be GPS observation sequence of length N, $Y = \{y_1, ..., y_{2N-1}\}$ be the label sequence and $t = 1..N$ be the position index in the sequence. The CRFs is formulated as follows:

$$P(Y|X) = \frac{1}{Z} \prod_{t=1}^{N} \exp(\sum_k \omega_k f_k(y_{2t-1}, x_t) + \sum_s \mu_s g_s(y_{2t}, y_{2t-1}, y_{2t+1}, X))$$

where $\{y_{2t-1}\}$ is the random variables over nearby roads of GPS observation $\{x_t\}$ called point node variables, $\{y_{2t}\}$ is the random variables over a finite set of feasible paths1 between two subsequent GPS observations called path node variables, $f_k$ and $g_s$ are the feature functions defined on the point nodes and path nodes respectively, $\omega_k$ and $\mu_s$ are the associated weights, and Z is the normalization term. This yields a chain-structured CRFs and a simplified example is illustrated in Figure 2. The map on top illustrates the simplified situation of identifying road states and path states given GPS observations in the road network. 5 random variables over the finite sets of road states and path states are respectively required $y_1:\{r_1, r_2\}$, $y_2:\{p_1, p_2, p_3\}$, $y_3:\{r_3, r_4\}$, $y_4:\{p_4, p_5\}$, $y_5:\{r_5, r_6\}$, to build the CRFs for map matching. Nodes $y_1, y_3, y_5$ linking with observations (black circles) are point nodes while nodes $y_2, y_4$ are path nodes. And a detailed description can be found in [10].

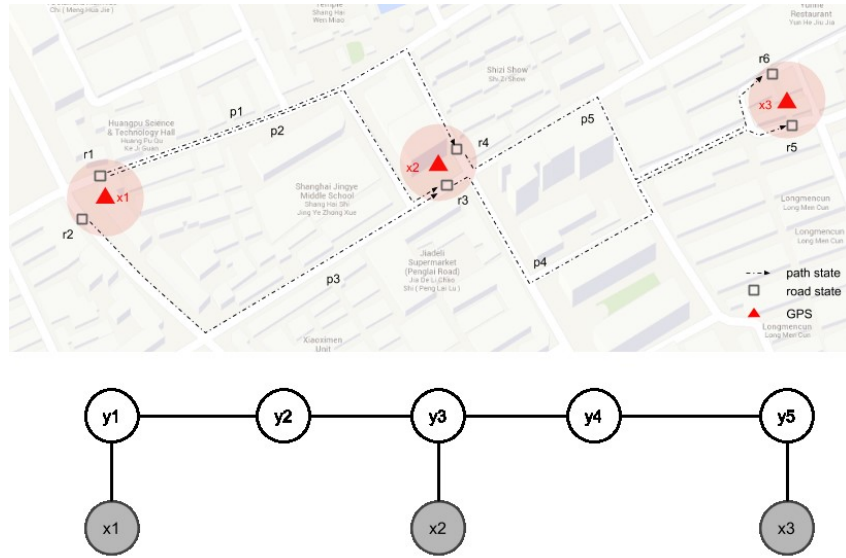

**Fig. 2.** A chain-structured CRFs for 3 GPS observations

---

[1] Feasible paths are those paths satisfy the routing constraints such as driving directions in the road network.

Then map matching can be fulfilled by performing inference on the CRFs. For a chain-structured CRFs, exact inference can be efficiently achieved using dynamic programming, e.g., Viterbi algorithm and an unconstrained optimization method can be used to estimate the weights of the features [11].

## 3  Feature Engineering for Map Matching

We construct two types of features, namely the point features and the path features for the chain-structured CRFs. The point features are built to discriminate the true road states given the noisy GPS observations while the path features are constructed to capture the routing choices between subsequent GPS observations in the road networks. For GPS trajectories at a high sample rate (sample interval shorter than 60s), using only 2 features, one for each type of node, would yield satisfying results. However, the performance drops dramatically for GPS trajectories at a low sample rate [12] which triggers the need of using more features.

### 3.1  Feature Construction with Geospatial Data

Unlike text and image data involved in other typical machine learning tasks, feature construction from vector-based geospatial data often requires to: 1) preprocess diverse data types, e.g., road networks are usually modeled using a graph model and GPS observations are treated as point objects in GIS; 2) explore the spatial relationship between spatial entities using spatial analysis, such as query the nearest point on the target road segment given GPS observation, compare the bearing of the road segment and GPS observation, etc. And this is more complex than in non-spatial database [13].

In the context of map matching, we have to deal with two types of geospatial data, GPS observations and road network. The raw GPS observations, stored in text files, are recorded from more than 7000 taxis traveling around in the urban road network of Shanghai, China, which results daily 20 million records (see Table 1. for an example, speed is recorded in km/h, direction is an integer ranging from 0 to 359, and occ. a binary attribute indicating if the taxi is occupied with passengers.), and the road network is extracted from OpenStreetMap (OSM) [14], a project that creates and distributes open geographic data for the world.

Table 1. An example record of raw GPS observations.

| car_id | longitude | latitude | speed | direction | occ. | timestamp |
|--------|-----------|----------|-------|-----------|------|-----------|
| 12971 | 121.360958 | 31.187778 | 61.2 | 249 | 1 | 2010-03-31 20:37:31 |

In order to efficiently store, inquire and analyze such a large volume of spatial data for feature construction, we develop the feature construction module on the top of a spatial database for its scalability to larger dataset and extensibility for more specialized functionality of spatial analysis. The processing pipeline of feature construction using a spatial database involves three steps:

1. *Data cleaning and conversion*. Before importing raw GPS text files and road network data, we remove the records with missing values or invalid values in the GPS text file (e.g., invalid timestamp and coordinates) and convert road data extracted from OSM to routable format, namely a graph-based representation, using osm2po[2].
2. *In-database feature construction*. Rather than moving the data in and out, we develop a server-side feature construction module to fulfill the task inside the database. More specifically, we implement the functions using the server programming language, PL/pgSQL, on PostgreSQL[3] with the spatial extension PostGIS[4] and the routing extension pgRouting[5] for basic spatial analysis and network analysis respectively. With the module, the feature construction can be simply performed as SQL query.
3. *Feature Scaling*. To avoid that certain features have dominant effects in the CRFs' loss function due to their relatively large values, features with continuous values are scaled to the range [0, 1].

### 3.2 Features for Map Matching

All though, the term of feature construction is not explicitly used in the literatures on map matching, the major modelling efforts of existing HMM-based map matching methods fall into this category. A brief review of mostly used features for map matching is given in [4].

**Table 2.** Example features for map matching

| feature | description | type |
| --- | --- | --- |
| distance error | distance between GPS observation to road state (closet point to the road segment) | point |
| bearing error | bearing difference between GPS heading and road direction | point |
| length | length of the path | path |
| maximum average speed | average speed limit of the road segments in the path | path |
| minimum average travel time | average travel time calculated using the speed limits of the road segments in the path | path |
| number of left turns | the count of left turns made in the path | path |
| number of right turns | the count of right turns made in the path | path |

Rather than using only a small set of features as previous HMM-based map matching methods do, we combine most informative features reported in the literatures [4][5] (some of the most used point and path features are listed in Table 2.). One of the challenges for feature construction from the large-scale taxi GPS trajectories is that the accuracy of the GPS devices are not well understood due to lack of specification of the positioning device. Thus, features built on the unreliable GPS observations may impose difficulty to identify the informative features in the training process. More specifically, we employ an accuracy filter for GPS observations at low speed since the GPS headings are reportedly unreliable at a low speed [15], $\text{val}\,I(v > v_{min}) + \text{val}_0 I(v < v_{min})$, where I is the indicator function, $\text{val}, \text{val}_0$ are the sensor reading and its specified initial value which are assigned to all road states when GPS speed v is lower than a specified threshold $v_{min}$. This filter is designed to capture the characteristics of the anonymous GPS and filter out the unreliable observations. And for path node features, we didn't use the feature *traffic_signals* and *stop_signs* because the related Point of Interest (POI) data is severely incomplete in the OSM dataset.

---
[2] http://osm2po.de
[3] http://www.postgresql.org
[4] http://postgis.net/
[5] http://pgrouting.org/

## 4   Preliminary Results

We test the model using all features from the literatures in comparison to the baseline methods on sample dataset drawn from Shanghai taxi GPS data. The sample dataset records one-day GPS trajectories of 70 taxis across the downtown area in Shanghai, China. It involves 124 trajectories in total and 13767 GPS observations covering an overall 788km. We manually labeled the sample dataset with a 10s-sample-interval and degrade it to obtain a sample interval of 120s using the even sampling strategy (this results in 1458 points and 1259 in-between paths). The dataset is split into a training set and a test set with a proportion of 7:3 on the trajectories, in which a portion of training set is used as a holdout set to tune the hyper parameter for regularization.

The preliminary results of the train/test error rates on both point and path nodes are obtained on the sample dataset and summarized in Table 3. The error rate is computed as the ratio of error predictions on individual points/paths to the total number of points/paths separately in order to examine the performance of point/path features accordingly. For the GPS dataset of 120 second sampling interval, the training set contains 1020 points and 881 paths while the test set contains 438 points and 378 paths. *base_complex* and *base_simple* are the baseline methods which use similar formalization of $\ell_2$ regularized CRFs taking only a small set of features [5]. *CRFs_L2* is $\ell_2$ regularized CRFs trained with BFGS [16], and *CRFs_L1* is $\ell_1$ regularized CRFs trained with Projected Scaled Sub-Gradient (PSS) methods [17]. The major differences compared to the baselines is that our methods combine a comprehensive feature set put forward in the literatures (including features used in baselines) and also employ a feature selection in the training phase using $\ell_1$ regularization. The results have shown that simply combining features from the literatures does not gain much improvement on the error rates. Furthermore, all methods suffer from a high training error rate on both point and path nodes, and training errors and test errors are relatively close. This suggests a high bias in the model. A potential improvement could be to extract more discriminating features of a path in the road network by incorporating context information, e.g., POIs which may help drivers remember the route along the path and at the turning point.

**Table 3.** Train/test error rates of map matching of GPS trajectories of 120s sample interval.

| Methods | #feature | Train Point | path | test point | Path |
|---|---|---|---|---|---|
| base_complex | 8 | .14 | .19 | **.15** | **.22** |
| base_simple | 2 | .14 | .19 | .17 | .26 |
| CRFs_L2 | 31 | **.11** | **.15** | .16 | .24 |
| CRFs_L1 | 8[6] | .13 | .17 | **.15** | **.22** |

In order to facilitate the future improvement, we also categorize the error instances (misclassified points and paths) in both training and test for CRFs_L1. The major error cases are *missing label* (18.3%), *parallel roads* (13.7%), *U-turn* (13.0%), *starting/ending point* (10.0%), and *position outlier* (9.9%). *Missing label* occurs when observations locate in the dense road networks and true states are unexpectedly eliminated due to the predefined count of states in implementation. *Parallel road* and *U-turn* happen when the model fits the observations well but makes no sense compared to real-world driving experience. *Starting/ending point* can be eliminated by combining contextual information, e.g., it's more likely to start a trip in the roads close to the building areas rather than in the middle of express roads. An overview of matching results and several error instances is illustrated in Figure 3.

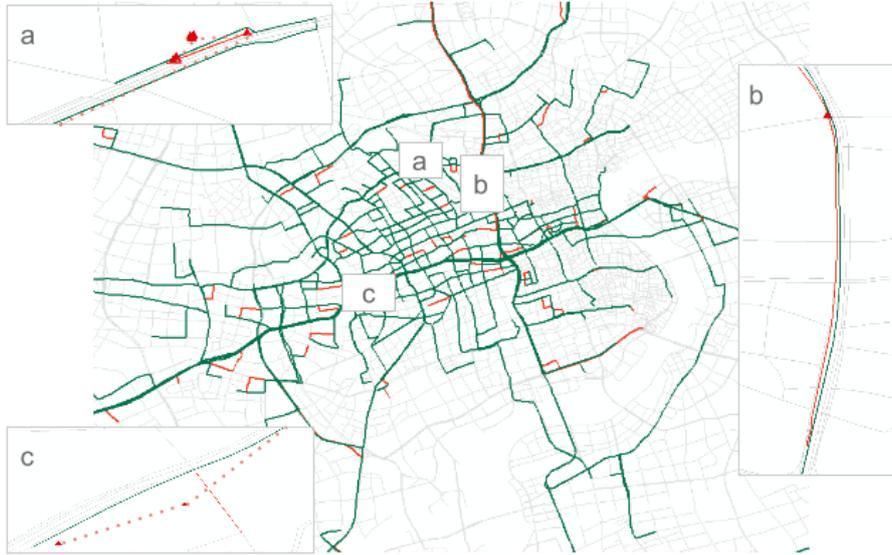

**Fig. 3.** Map matching results of CRFs_L1(red) overlaid by ground truth(green). a) starting/ending point, b) parallel roads, c) U-turn.

---

[6] CRFs_L1 is trained with 31 features same as CRFs_L2 while yielding 8 features out of the $\ell_1$ regularization.

## 5 Conclusion

In this work in progress, we report our recent work on feature engineering for map matching and specify two major uncertainty sources in modeling trajectory data in a road network, namely noisy sensor data and dynamic routing choices. We also share our experience on extracting features from a spatial database, which can serve as an example for other machine learning applications driven by spatial data. The preliminary result on sample data set from real-world taxi GPS trajectories indicates a high bias in the model and more discriminating features are needed.

## 6 Acknowledgement

The first author is financially supported by China Scholarship Council (CSC).